\documentclass[11pt]{article}
\usepackage{amssymb}
\usepackage{amsmath}
\usepackage[final]{acl}
\usepackage{subcaption}
\usepackage{times}
\usepackage{latexsym}
\usepackage{algorithm}
\usepackage{algorithmic}
\usepackage{CJKutf8}
\usepackage{newfloat}
\usepackage{listings}
\usepackage{pifont}
\usepackage{colortbl} 
\usepackage{array}
\usepackage{booktabs}
\usepackage{adjustbox}
\usepackage{tabularx} 
\usepackage{xcolor}
\usepackage{listings}
\usepackage{multirow}
\newcommand{\RNum}[1]{\uppercase\expandafter{\romannumeral #1\relax}}
\definecolor{myblue}{RGB}{212, 239, 251}
\definecolor{mygreen}{RGB}{217, 255, 217}
\definecolor{myyellow}{RGB}{255, 252, 200}
\newcommand{\Pmodal}{\ding{52} & }
\newcommand{\Rmodal}{ & \ding{52} }
\newcommand{\RPmodal}{\ding{52} & \ding{52}}
\newcommand{\modelname}{SignThought}
\newcommand{\corr}[1]{\textcolor{green!50!black}{#1}}          
\newcommand{\para}[1]{\textcolor{blue}{#1}} 
\newcommand{\wrong}[1]{\textcolor{red}{#1}}          
\usepackage[T1]{fontenc}

\usepackage[utf8]{inputenc}
\usepackage{color}

\usepackage{microtype}

\usepackage{inconsolata}

\usepackage{graphicx}
\usepackage{pifont}
\newcommand{\xmark}{\ding{55}} 
\title{Think in Latent Thoughts: A New Paradigm for Gloss-Free Sign Language Translation}

\author{
    \textbf{Yiyang Jiang}$^1$\quad \textbf{Li Zhang}$^1$\quad  \textbf{Xiaoyong Wei}$^{1,2}$\thanks{\ Corresponding author}\quad \textbf{Qing Li}$^1$ \\
    $^1$The Hong Kong Polytechnic University \quad $^2$Sichuan University\\
    \texttt{yiyang.jiang@connect.polyu.hk}\\
    \texttt{\{zanly20.zhang, cs007.wei, qing-prof.li\}@polyu.edu.hk}
}

\begin{document}
\maketitle

\begin{abstract}
%
Many Sign language translation (SLT) systems quietly assume that brief chunks of signing map directly to spoken-language words. 
That assumption breaks down because signers often create meaning on the fly using context, space, and movement.
We revisit SLT and argue that it is mainly a cross-modal reasoning task, not just a straightforward video-to-text conversion. 
We thus introduce a reasoning-driven SLT framework that uses an ordered sequence of latent thoughts as an explicit middle layer between the video and the generated text. 
These latent thoughts gradually extract and organize meaning over time. On top of this, we use a plan-then-ground decoding method: the model first decides what it wants to say, and then looks back at the video to find the evidence. 
This separation improves coherence and faithfulness.
We also built and released a new large-scale gloss-free SLT dataset with stronger context dependencies and more realistic meanings. 
Experiments across several benchmarks show consistent gains over existing gloss-free methods. Our code and data are available at \url{https://github.com/fletcherjiang/SignThought}.
%
\end{abstract}

\section{Introduction}
Sign language translation (SLT) is critically important both as a vital assistive technology for connecting Deaf and hard-of-hearing communities and as a challenging multimodal task within natural language processing~\cite{bragg2019interdisciplinary}. 
Research in this area has evolved significantly, progressing from early methods that approached SLT as a gloss-level classification problem over video segments~\cite{koller2015continuous} to more recent formulations as a gloss-free, video-to-text sequence transduction task~\cite{lin2023gloss,chen2024c2rl}, tackled by specialized architectures or multimodal large language models~\cite{wong2024sign2gpt,chen-etal-2024-fla-llm,gong2024signllm}. 
While these efforts have yielded encouraging outcomes, substantial room for improvement remains.

\begin{figure}[t!]
    \centering
    \includegraphics[width=1\linewidth]{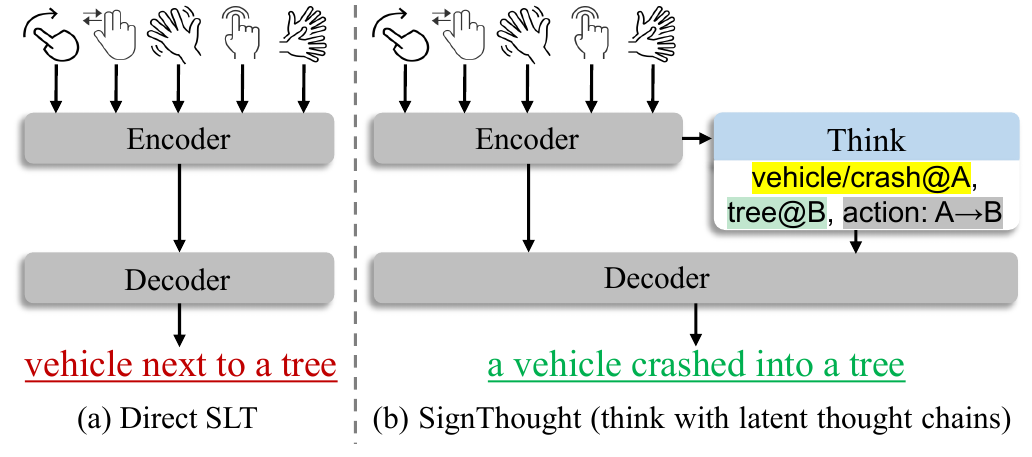}
    \caption{Latent thoughts clarified the context to disambiguate signs and guide decoding toward the correct object interaction in gloss-free SLT.}
    \label{fig:introlatent}
    \vspace{-1.5em}
\end{figure}

A key limitation lies in a long-standing misconception in model design: treating sign video segments as if they were directly mappable to static spoken-language words or phrases~\cite{yin2021including,camgoz2018neural,bragg2019interdisciplinary}. 
This view originates from an early emphasis on the \textit{Frozen Lexicon}, i.e., the finite inventory of conventionalized signs listed in glossaries, while largely overlooking \textit{Productive Forms}~\cite{forster2014extensions}. 
Productive forms are constructed on-the-fly through classifiers, spatial grammar, and motion modulation, forming an open-ended and context-governed system that cannot be enumerated as discrete lexical entries~\cite{zwitserlood2012classifiers}. 
For instance, a single ``vehicle'' handshape may express meanings such as ``park,'' ``crash,'' or ``drive'' purely through variations in movement and spatial configuration~\cite{zwitserlood2012classifiers,chongwahngo2008trecvid}. 
As a consequence, meaning in sign language is often not explicitly encoded in fixed symbolic units but dynamically \emph{generated} from motion, space, and discourse context~\cite{zwitserlood2012classifiers,cormier2012lexicalisation,WXY_CAL_TIP_2013}. SLT is therefore a generative, context-dependent reasoning problem, not just segment alignment or symbol substitution ~\cite{yin2021including,bragg2019interdisciplinary,gong2024signllm,jiang2024prior}. 


From this view, SLT naturally aligns with recent advances in {text reasoning} techniques, where models explicitly maintain intermediate semantic states to support multi-step meaning composition under weak or implicit supervision~\cite{wei2022cot,yao2022react}.
However, unlike text-only reasoning, SLT reasoning must operate across heterogeneous visual and linguistic modalities~\cite{mitra2024compositional}. 
The absence of discrete reasoning primitives and explicit intermediate representations makes existing reasoning techniques difficult to transfer directly, leaving multimodal reasoning in SLT largely unexplored~\cite{yin2021including}. 
We therefore position this work as a {pilot study} toward understanding how explicit reasoning can be modeled in gloss-free SLT, with the goal of identifying fundamental challenges and exploring principled solutions~\cite{bragg2019interdisciplinary,jiang2025self}. 
%
This paradigm offers three key advantages that tackle the reasoning challenges as:  
\begin{itemize}
    \item \textbf{Latent Thought Abstraction.} We introduce latent thoughts as an explicit intermediate semantic interface between sign videos and text generation. Rather than compressing all semantic information into opaque encoder features, the model maintains an ordered set of latent reasoning states that progressively distill and organize meaning from long, continuous visual streams, thereby bridging the mismatch between continuous visual evidence and discrete reasoning primitives.
    \item \textbf{Plan–Ground Decoupling.} Reasoning and grounding are explicitly separated through a plan-then-ground generation mechanism: the model first determines \emph{what} semantic content to express by reasoning over latent thoughts, and only then decides \emph{where} to retrieve the corresponding visual evidence. This decoupling alleviates the strong entanglement between semantic decision-making and evidence retrieval in existing SLT systems, improving controllability and grounding faithfulness.
    \item \textbf{Traceable Evidence Alignment.} Latent thoughts function not only as internal planning states but also as traceable anchors that align generated text with specific temporal regions of the input video, enabling explicit evidence attribution and more faithful translations, while a newly constructed large-scale gloss-free SLT dataset provides the essential empirical basis for studying and validating such reasoning behaviors. 

\end{itemize}


\section{Related Work}
\paragraph{Sign Language Recognition and Translation.}
Sign Language Recognition (SLR) maps sign videos to gloss sequences, evolving from isolated (ISLR)~\cite{albanie2020bsl,li2020word} to continuous settings (CSLR), with typical pipelines extracting visual features from RGB or pose-based inputs~\cite{chen2022two}, modeling temporal dynamics via RNN/LSTM or Transformer architectures~\cite{camgoz2018neural,camgoz2020sign}, and decoding glosses using HMMs~\cite{8099847} or CTC objectives~\cite{cheng2020fully}. While effective, such systems are computationally intensive and limited in exploiting higher-level linguistic context. Sign Language Translation (SLT) further translates sign videos into spoken or written language, where most methods remain gloss-based by cascading SLR with translation or jointly optimizing both tasks~\cite{camgoz2020sign,zhou2021improving}. In contrast, gloss-free SLT directly learns video-to-text mappings using Transformer-based or variational models~\cite{li2020tspnet,camgoz2018neural,tu2026easlt}; although recent advances leveraging large-scale multimodal pretraining~\cite{li2025uni} and large language model adaptation~\cite{wong2024sign2gpt} improve translation fluency, they introduce substantial computational overhead and increased dependence on external corpora.

\begin{figure*}[htbp]
    \centering
    \includegraphics[width=1\textwidth]{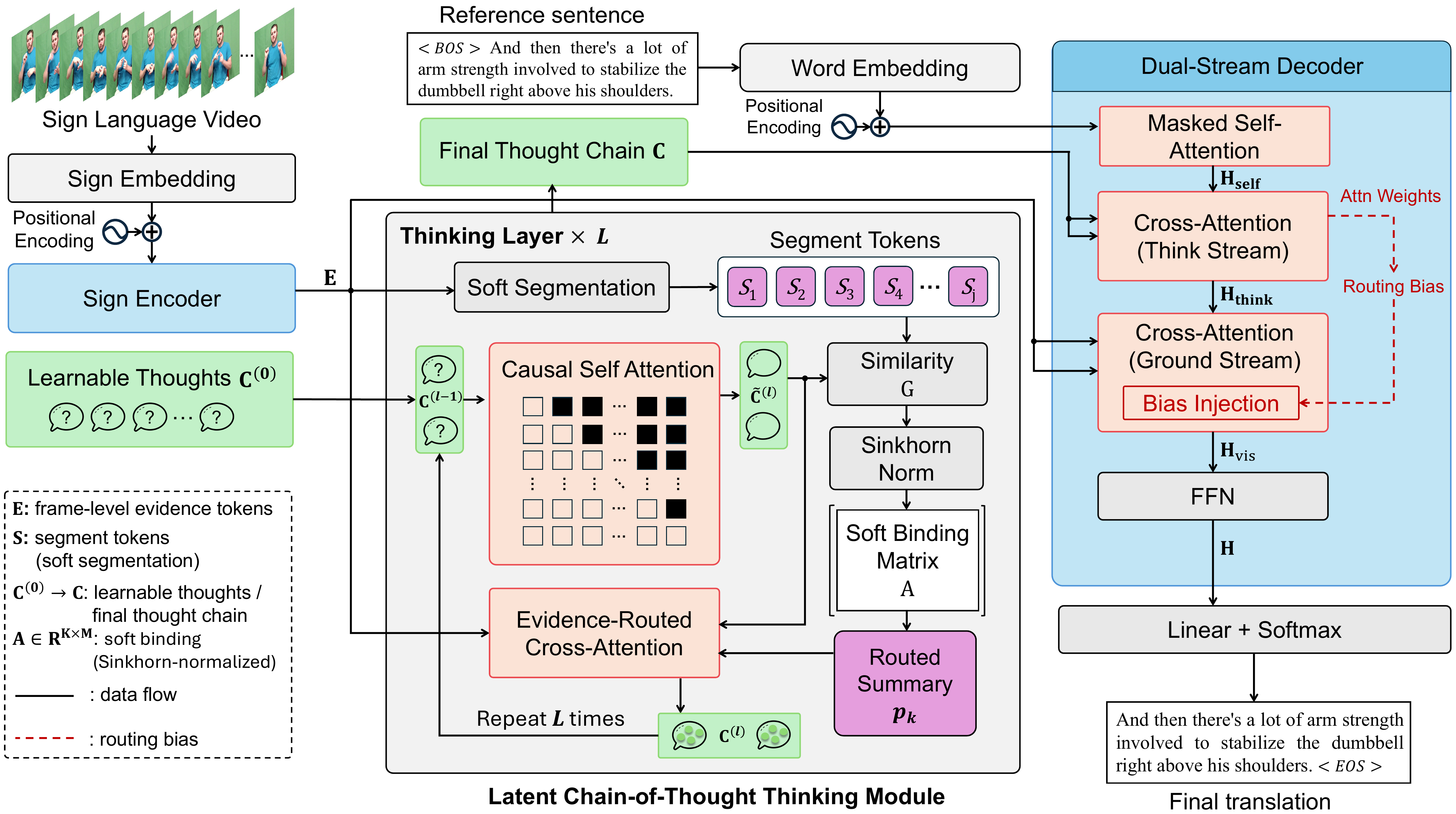}
    \caption{An overview of the \modelname~framework, which consists of three parts: (i) a sign encoder that maps the input video $\mathbf{X}$ into dense evidence features $\mathbf{E}$; (ii) a Latent Chain-of-Thought module that iteratively updates $K$ learnable thought slots and performs structured routing (segmentation $\rightarrow$ Sinkhorn-style binding $\rightarrow$ routed retrieval, with an optional monotonic bias) to distill $\mathbf{E}$ into an ordered thought chain $\mathbf{C}$; and (iii) a plan-then-ground dual-stream decoder that first attends to $\mathbf{C}$ for semantic planning and then grounds each token on $\mathbf{E}$ under thought-guided temporal priors to generate the spoken-language translation end-to-end.}
    \label{fig:pipeline}
    \vspace{-1.0em}
\end{figure*}

\paragraph{Reasoning and Latent Thoughts.}
Chain-of-Thought (CoT) prompting enhances multi-step reasoning in large language models (LLMs) by eliciting intermediate reasoning steps~\cite{wei2023chainofthoughtpromptingelicitsreasoning,zhou2024enhancing,zhou2025valuing}, with extensions such as self-consistency~\cite{wang2023selfconsistencyimproveschainthought}, planning, search, and tool-augmented reasoning further improving robustness and capability~\cite{yao2023reactsynergizingreasoningacting,xie2025unlocking,zeng2025FSDrive}. Beyond general reasoning tasks, LLMs have also been applied to personalized recommendation assistants with memory and reasoning~\cite{huang2026towards,huang2025mr}. CoT has also been explored in multimodal grounding and video reasoning~\cite{zhang2024multimodalchainofthoughtreasoninglanguage}. More recently, reasoning has shifted from discrete language tokens to continuous latent spaces, where hidden states are reused as “latent thoughts” through feedback or iterative computation~\cite{hao2025traininglargelanguagemodels}, with formal analyses and surveys clarifying the relationship between CoT and latent-thought paradigms~\cite{xu2025formalcomparisonchainofthoughtlatent}. Notably, existing latent-thought approaches primarily operate on token-level embeddings within unimodal LLMs, whereas our Latent Thoughts are video-conditioned latent slots updated via attention over sign-video features, explicitly serving as a cross-modal interface bridging visual evidence and decoding in gloss-free SLT.

\section{Method}
We study gloss-free sign language translation (SLT), where the supervision consists solely of sentence-level translations without any intermediate gloss annotations. Given a sign video clip, we aim to generate a target spoken-language sentence. Let a sign clip be represented as a sequence of visual observations
$\mathbf{X}=\{x_t\}_{t=1}^{T_s}$ (pre-extracted features).
A binary mask $\mathbf{m}\in\{0,1\}^{T_s}$ indicates valid time steps when clips are padded. The target translation is a token sequence
$\mathbf{Y}=\{y_t\}_{t=1}^{T_t}$ with $y_t\in\mathcal{V}$, where $\mathcal{V}$ is the text vocabulary. We assume a dataset of paired samples
$\mathcal{D}=\{(\mathbf{X}^{(n)}, \mathbf{Y}^{(n)})\}_{n=1}^{N}$.

Crucially, no gloss sequence or sign-token alignment is provided. We learn a conditional sequence model that maximizes the likelihood of the target sentence given the sign clip:
\begin{equation}
p_\theta(\mathbf{Y}\mid \mathbf{X})
=\prod_{t=1}^{T_t} p_\theta\!\left(y_t \mid y_{<t}, \mathbf{X}\right).
\end{equation}
In addition to the standard autoregressive formulation, our method introduces a set of $K$ ordered latent thought states $\mathbf{C}=\{c_k\}_{k=1}^{K}$ that serve as an intermediate semantic plan distilled from $\mathbf{X}$ and consumed by the decoder during generation.

\subsection{Sign Encoder}
\label{sec:encoder}
Given a sign video clip, we represent it as a feature sequence
$\mathbf{X}=\{x_t\}_{t=1}^{T_s}$, where each $x_t\in\mathbb{R}^{d_x}$ is a pre-extracted visual feature vector from a pretrained Inception network at time step $t$, and $T_s$ is the padded sequence length. Following the feature setting of~\cite{voskou2021stochastic}, we initialize the visual feature extractor from the open-sourced Inception network and remove gloss-dependent supervision during feature learning. The extractor is trained with a sentence-level contrastive objective using only paired sign video-sentence data. We use a binary mask $\mathbf{m}\in\{0,1\}^{T_s}$ to indicate valid (non-padded) positions. We first project the raw per-frame features into the model space via a learnable embedding layer, i.e., $\mathbf{E}^{(0)} = \mathrm{Embed}(\mathbf{X})$,
where $\mathrm{Embed}(\cdot)$ is implemented as a linear projection followed by normalization and dropout. This step unifies sign features into a shared representation space for subsequent sequence modeling.

To inject order information, we add positional encoding to obtain $\hat{\mathbf{E}}^{(0)}$, with dropout applied afterward. We then stack $N_{\text{enc}}$ encoder layers to model long-range dependencies and local motion patterns:
\begin{equation}
\hat{\mathbf{E}}^{(n)} = \mathrm{EncBlock}^{(n)}(\hat{\mathbf{E}}^{(n-1)}, \mathbf{m}), \ \ n=1,\dots,N,
\end{equation}
where each $\mathrm{EncBlock}$ follows a Conformer-style design, combining multi-head self-attention (masked by $\mathbf{m}$) with a lightweight convolutional module to capture local temporal dynamics, together with feed-forward sublayers and residual normalization. The final encoder output is $\mathbf{E}=\hat{\mathbf{E}}^{(N)}=\{e_t\}_{t=1}^{T_s}.$
The resulting sequence $\mathbf{E}$ serves as the frame-level evidence source for both the Latent CoT thinking module (Sec.~\ref{sec:thinking}) and the dual-stream decoder (Sec.~\ref{sec:decoder}).

\subsection{Thinking in Latent Chain-of-Thought}
\label{sec:thinking}
The objective of the thinking module is to transform the dense encoder representations $\mathbf{E}$ into a compact and ordered set of $K$ latent thought states $\mathbf{C}=\{c_k\}_{k=1}^{K}$, which collectively serve as a high-level semantic plan for translation. In contrast to gloss-based pipelines, these thought states are learned end-to-end under sentence-level supervision and are explicitly organized to form a directional latent chain. Given encoder features $\mathbf{E}=\{e_t\}_{t=1}^{T_s}$ and corresponding source mask, the module outputs $\mathbf{C} = \mathrm{Think}(\mathbf{E}) \in \mathbb{R}^{K\times d}$, where the index $k$ defines the causal order of the latent reasoning process.

\paragraph{Learnable thought slots.}
A central difficulty in gloss-free SLT is that the video provides dense and redundant evidence, while the translation requires a compact sequence of semantic decisions. Instead of forcing the decoder to discover this structure on the fly, we introduce an explicit intermediate interface: a small set of ordered thought slots that serve as the model’s working memory for step-wise semantic composition.
Concretely, we initialize the thought chain with $K$ learnable slots $\mathbf{C}^{(0)}=\{c_k^{(0)}\}_{k=1}^{K}$, where each $c_k^{(0)}\in\mathbb{R}^{d}$ is a trainable parameter, serving as empty reasoning states before evidence injection. These slots can be viewed as \emph{empty} reasoning states that will be filled with semantic evidence distilled from $\mathbf{E}$.

\paragraph{Thinking layers.}
A key bottleneck in gloss-free SLT is that the encoder produces a long, continuous evidence stream $\mathbf{E}$, while what we ultimately need for generation is a compact and structured semantic plan.
If we let the decoder directly attend to all frames, it must simultaneously decide what to express and where the supporting evidence lies, which often leads to diffuse attention and unstable evidence tracking.
We therefore introduce a dedicated thinking stage that repeatedly compresses and reorganizes $\mathbf{E}$ into an ordered latent chain $\mathbf{C}$ before decoding. To make this compression process expressive yet controllable,
we stack $L$ identical thinking layers to iteratively refine the chain:
\begin{equation}
\mathbf{C}^{(\ell)} = \mathrm{ThinkLayer}^{(\ell)}(\mathbf{C}^{(\ell-1)}, \mathbf{E}), \ \ell=1,\dots,L,
\end{equation}
and set $\mathbf{C}=\mathbf{C}^{(L)}$ as the final latent chain-of-thought. Each thinking layer consists of two main steps: causal thought mixing to impose an ordered chain prior, and an evidence-routed cross-attention operator that retrieves sign evidence in a structured manner.

We first encourage a directional (coarse-to-fine) refinement inside the latent chain.
Specifically, we apply masked self-attention over the $K$ thought states using a lower-triangular (causal) mask, so that the $k$-th thought only attends to $\{1,\dots,k\}$:
\begin{equation}
\tilde{\mathbf{C}}^{(\ell)} = \mathrm{CausalSelfAttn}\!\left(\mathbf{C}^{(\ell-1)}\right).
\end{equation}
This simple prior turns the thought set into an ordered workspace: earlier thoughts summarize coarse semantics and provide context, while later thoughts refine with additional details, making the intermediate reasoning process more stable and interpretable.

The second challenge is evidence assignment. A standard cross-attention from thoughts to all frames is both inefficient on long clips and prone to degeneracy, where multiple thoughts attend to the same frames. We address this by routing before retrieval: each thought is first encouraged to be approximately contiguous (via Lcont), and only then does it retrieve fine-grained evidence within that responsibility region.

From the frame-level stream $\mathbf{E}$, we derive (i) frame evidence (the original tokens), (ii) a small set of segment-level tokens $\mathbf{S}$ obtained by a differentiable soft segmentation over time~\cite{OFFSET}. Intuitively, $\mathbf{S}$ provides sparse semantic ``chunks'' (analogous to latent phrase units) while frame evidence preserves fine-grained lexical details. We compute segment-to-frame weights $\mathbf W_{seg}\in\mathbb{R}^{M\times T_s}$ via soft boundaries (per instance), where each row defines a soft temporal window and is normalized over frames.
Segment tokens are obtained by weighted pooling:
\begin{equation}
S_j=\sum_{t=1}^{T_s}(\mathbf W_{seg})_{j,t}\,e_t,\qquad j=1,\ldots,M.
\end{equation}

Given segment tokens $\mathbf{S}=\{S_j\}_{j=1}^{M}$ and latent thoughts $\mathbf{C}=\{c_k\}_{k=1}^{K}$, we compute a similarity score matrix $\mathbf{G}\in\mathbb{R}^{K\times M}$ with
\begin{equation}
G_{k,j} = \frac{(W_q c_k)^\top (W_k S_j)}{\sqrt{d}} .
\end{equation}
We then apply a differentiable Sinkhorn-style normalization to obtain a soft binding matrix
$\mathbf{A}=\mathrm{Sinkhorn}(\mathbf{G})\in\mathbb{R}^{K\times M}$, where $A_{k,j}\ge 0$ denotes the (soft) amount of segment $j$ assigned to thought $k$.
In practice, we perform Sinkhorn normalization with a \emph{row-wise unit mass} and a \emph{uniform column budget}, i.e.,
$\sum_{j=1}^{M} A_{k,j}\approx 1$ for all $k$ and $\sum_{k=1}^{K} A_{k,j}\approx K/M$ for all $j$ (up to a small error $\epsilon_{\mathrm{num}}$).
This yields an approximately balanced transport plan: each thought receives comparable total evidence, while each segment contributes a limited and evenly distributed amount across thoughts, preventing collapsed routing while keeping the assignment fully soft. Each thought receives a routed segment summary:
\begin{equation}
p_k = \sum_{j=1}^{M} A_{k,j}\, S_j,\qquad p_k\in\mathbb{R}^{d},
\end{equation}
where $p_k$ aggregates the segment evidence assigned to thought $c_k$. Since $\mathbf{A}$ is approximately row-normalized (i.e., $\sum_{j}A_{k,j}\approx 1$), $p_k$ can be interpreted as a weighted average of segment tokens.

Conditioned on the routed summary $p_k$, each thought retrieves fine-grained sign evidence from $\mathbf{E}$ via cross-attention (and optionally sparse/deformable retrieval for efficiency on long clips). We denote the resulting evidence-aware thought update by
\begin{equation}
\hat{\mathbf{C}}^{(\ell)} = \mathrm{RoutedXAttn}\!\left(\tilde{\mathbf{C}}^{(\ell)}, \mathbf{E}, \mathbf{p}\right),
\mathbf p=\{p_k\}_{k=1}^K
\end{equation}
where $\mathrm{RoutedXAttn}$ encapsulates \emph{(evidence-fabric $\rightarrow$ routing $\rightarrow$ retrieval)} as a single operator in our implementation and in the overview figure, and uses $p_k$ (equivalently, the routing induced by $\mathbf{A}$) to bias the cross-attention over $\mathbf{E}$. Finally, we apply a position-wise feed-forward network with residual normalization to produce the layer output:
\begin{equation}
\mathbf{C}^{(\ell)} = \mathrm{FFN}\!\left(\hat{\mathbf{C}}^{(\ell)}\right).
\end{equation}

The final thought chain $\mathbf{C}$ is consumed by the decoder through thinking cross-attention (Sec.~\ref{sec:decoder}). In addition, the intermediate routing variables (e.g., the thought-to-segment binding $\mathbf{A}$ and the segmentation weights) can be cached as an internal prior to guide visual grounding in the dual-stream decoder, without changing the model Input and Output.

\subsection{Dual-Stream Decoder}
\label{sec:decoder}
We adopt an autoregressive Transformer decoder that explicitly follows a \emph{plan-then-ground} generation pattern. 
At each layer, the decoder first consults the latent thought chain to decide \emph{what} semantic content to express, and then returns to the encoder features to retrieve fine-grained signing evidence for \emph{how} to realize that content in fluent text.

Given an input text prefix $\mathbf{y}_{<t}$ (teacher forcing during training), we embed tokens with standard word and positional (and optionally token-type) embeddings to obtain decoder states.

Let $\mathbf{H}^{(l-1)}$ denote the decoder hidden states entering layer $l$. Each layer contains the following sub-layers. We first apply masked self-attention over previously generated tokens to produce $\mathbf{H}^{(l)}_{\mathrm{self}}=\mathrm{MSA}\!\left(\mathbf{H}^{(l-1)}\right)$, where the subsequent mask ensures autoregressive generation.

We then attend to the latent thought chain $\mathbf{C}$:
\begin{equation}
\mathbf{H}^{(l)}_{\mathrm{think}}=\mathrm{XAttn}\!\left(\mathbf{H}^{(l)}_{\mathrm{self}},\,\mathbf{C}\right).
\end{equation}
Intuitively, this step selects the current semantic ``reasoning state'' from the ordered thoughts, providing a high-level plan for the next tokens. We also retain the corresponding attention weights over thoughts,
denoted by $\boldsymbol{\alpha}$, as they reflect which thought(s) each target position relies on.

To make ``think first, then ground'' a mechanism rather than only an ordering, we derive a soft temporal retrieval prior from $\boldsymbol{\alpha}$ using intermediate routing variables produced by the thinking module (per instance). Specifically, we reuse (i) the thought-to-segment binding matrix $\mathbf{A}\in\mathbb{R}^{K\times M}$, and (ii) the segment-to-frame aggregation weights $\mathbf{W}_{seg}\in\mathbb{R}^{M\times T_s}$. We map token-to-thought attention to a token-to-frame prior:
\begin{equation}
\boldsymbol{\beta} = \boldsymbol{\alpha}\mathbf{A},\qquad
\mathbf{w} = \boldsymbol{\beta}\mathbf{W}_{seg},
\end{equation}
where $\boldsymbol{\alpha}\in\mathbb{R}^{T_t\times K}$ denotes the head-averaged \emph{attention weights} over $K$ thoughts for each target position, and $\mathbf{w}\in\mathbb{R}^{T_t\times T_s}$ assigns each target position a soft preference over source time steps.

Here $\mathbf{A}$ is approximately row-normalized over segments (i.e., $\sum_{j=1}^{M}A_{k,j}\approx 1$) and $\mathbf{W}_{seg}$ is row-normalized over frames (i.e., $\sum_{t=1}^{T_s}(\mathbf{W}_{seg})_{j,t}=1$), with all entries non-negative, so that each row of $\mathbf{w}$ forms an (approximately) normalized soft temporal prior. This routing prior is internal (no change to model I/O) and can be omitted without affecting the decoder interface. Finally, we attend to the encoder features $\mathbf{E}$ to obtain grounded decoder states:
\begin{equation}
\mathbf{H}^{(l)}_{\mathrm{vis}}=\mathrm{XAttn}\!\left(\mathbf{H}^{(l)}_{\mathrm{think}},\,\mathbf{E}\right),
\end{equation}
with the standard source mask applied. When the routing prior $\mathbf{w}$ is enabled, we incorporate it as a soft additive bias in the attention logits, encouraging the decoder to retrieve evidence from time regions consistent with the selected thought(s). This yields thought-guided evidence retrieval while preserving the standard Transformer decoder structure~\cite{ReTrack,REFINE}.



A position-wise FFN with residual connections and layer normalization produces the layer output: $\mathbf{H}^{(l)}=\mathrm{FFN}\!\left(\mathbf{H}^{(l)}_{\mathrm{vis}}\right).$
The final decoder states are projected to the vocabulary to obtain token probabilities, and the model is trained with the standard sequence-level cross-entropy objective (Sec.~\ref{sec:objective}).


\begin{table*}[htbp]
  \centering
    \small
    \setlength{\tabcolsep}{5pt}
    \resizebox{1\linewidth}{!}{
    \begin{tabular}{l|cc|ccccc|ccccc}
       
    \toprule
    { \multirow{2}{*}{\textbf{Method}}} & \multicolumn{2}{c|}{\textbf{Modality}} & \multicolumn{5}{c|}{\textbf{PHOENIX14T}}& \multicolumn{5}{c}{\textbf{CSL-Daily}}\\
    \cmidrule(lr){2-3}  \cmidrule(lr){4-8} \cmidrule(lr){9-13} 
    & Pose & RGB & B@1  & B@2 & B@3 & B@4 & ROUGE & B@1  & B@2 & B@3 & B@4 & ROUGE \\
    \midrule
        \rowcolor[gray]{.85} \multicolumn{13}{c}{Gloss-based}\\
    \midrule

    SLRT~\citep{camgoz2020sign} &  \Rmodal & 46.61 & 33.73 & 26.19 & 21.32 & - & 37.38 &  24.36 &  16.55 &  11.79 &  36.74   \\
    
    
    SignBT~\citep{zhou2021improving} &  \Rmodal & 50.80 & 37.75 & 29.72 & 24.32 & 49.54 &  51.42 &  37.26 &  27.76 &  21.34 &  49.31  \\
    
    MMTLB~\citep{chen2022simple}&  \Rmodal & 53.97 &41.75 &33.84 &28.39 &52.65 &  
    53.31 &  40.41 &  30.87 &  23.92 &  53.25  \\

     SLTUNET~\citep{zhang2023sltunet} &  \Rmodal & 52.92 &41.76 &33.99 &28.47 &52.11 &  54.98 & 41.44 & 31.84 & 25.01 & 54.08   \\
    
    CV-SLT~\citep{zhao2024conditional} &  \Rmodal &   54.88 & 42.68 & 34.79 & 29.27 & 54.33 &  {58.29} & {45.15} & {35.77} & {28.94} & {57.06}  \\
    
    TS-SLT~\citep{chen2022two} &  \RPmodal & 54.90 &42.43 &34.46 &28.95 &53.48 &  55.44 &  42.59 &  32.87 &  25.79 &  55.72  \\

    \midrule
        \rowcolor[gray]{.85} \multicolumn{13}{c}{Weakly supervised gloss-free}\\
    \midrule
     GASLT~\citep{yin2023gloss} &  \Rmodal &  39.07& 26.74& 21.86& 15.74&39.86 & 19.90 &  9.94 &  5.98 &  4.07 &  20.35 \\
     
     {VAP}~\citep{jiao2024visual}  & \Pmodal & {53.07} & {-} & {-} & {26.16} & {51.28} & {{52.98}} & {{-}} & {{-}} &  {{23.65}} & {{51.09}} \\

    \midrule
        \rowcolor[gray]{.85} \multicolumn{13}{c}{Gloss-free}\\
    \midrule

    NSLT~\citep{camgoz2018neural} &  \Rmodal &  32.24 & 19.03 & 12.83 & 9.58 & 31.80 & 34.16 &  19.57 &  11.84 &  7.56 &  34.54  \\

    GFSLT-VLP~\citep{zhou2023gloss} &  \Rmodal & 43.71 &33.18 &26.11 &21.44 &42.29 &  39.37 &  24.93 &  16.26 &  11.00 &  36.44  \\

    MSLU~\citep{zhou2024scaling} &  \Pmodal&  46.56 & 34.21 & - & 22.24 & 46.73 & 33.97 & 22.20  & -  &  11.42 & 33.80\\
    
    FLa-LLM~\citep{chen-etal-2024-fla-llm} &  \Rmodal &46.29& 35.33& 28.03& 23.09& 45.27 & 37.13 & 25.12 &  18.38 &   14.20 &  37.25  \\
    
    Sign2GPT~\citep{wong2024sign2gpt} &  \Rmodal &  {49.54} &{35.96}& {28.83}& 22.52& {48.90} & {41.75} & {28.73} & {20.60} &  15.40 & {42.36} \\
    SignLLM~\citep{gong2024signllm} &  \Rmodal & 45.21 & 34.78 & 28.05 &{23.40} &44.49 & {39.55} & {28.13} & {20.07} & {15.75} & {39.91} \\
    {C$^2$RL}~\citep{chen2024c2rl}  & \Rmodal & {52.81} & \underline{40.20} & \textbf{32.20} & {26.75} & {50.96} & \underline{{49.32}} & \underline{{36.28}} & \underline{{27.54}} &  \underline{{21.61}} & \underline{{48.21}} \\
    \midrule
    \textbf{\modelname~(Ours)} &  \Rmodal& \underline{51.18}  & \underline{39.40} & \underline{32.17} & \textbf{27.22} & \textbf{54.50} & \textbf{49.57}  & \textbf{37.90} & \textbf{29.33}
    & \textbf{23.92} & \textbf{50.99} \\
    \bottomrule
    \end{tabular}}
    \vspace{-0.7em}
    \caption{SLT results on PHOENIX14T and CSL-Daily datasets. The best results are highlighted as \textbf{bold}, and the second-best are \underline{underlined}.}
    \label{tab:sota}
    \vspace{-0.8em}
\end{table*}

\subsection{Training Objective}
\label{sec:objective}


\paragraph{Translation loss.}
Given a mini-batch of paired data $\{(\mathbf{X}^{(b)},\mathbf{Y}^{(b)})\}_{b=1}^{B}$,
where $B$ is the batch size and $T_t^{(b)}$ denotes the target length of the $b$-th sample,
we train the dual-stream decoder with teacher forcing and standard cross-entropy:
\begin{equation}
\mathcal{L}_{ce}
= -\frac{1}{B}\sum_{b=1}^{B}\sum_{t=1}^{T_t^{(b)}}
\log p_\theta\!\left(y^{(b)}_t \mid y^{(b)}_{<t}, \mathbf{E}^{(b)}, \mathbf{C}^{(b)}\right).
\end{equation}

\paragraph{Latent chain regularization.}
To encourage the latent thoughts to behave as an ordered chain of contiguous semantic units, we add two simple regularizers on the thought-to-segment assignment matrix $\mathbf{A}$ produced by the thinking module (Sec.~\ref{sec:thinking}).

Let $\mu_k$ be the expected segment index assigned to the $k$-th thought:
\begin{equation}
\mu_k^{(b)}=\sum_{j=1}^M j\cdot A_{k,j}^{(b)}
\end{equation}
We penalize violations of the forward (coarse-to-fine) progression:
\begin{equation}
\mathcal{L}_{mono}
=\frac1B\sum_{b=1}^B\sum_{k=1}^{K-1}\mathrm{ReLU}(\mu_k^{(b)}-\mu_{k+1}^{(b)}+\delta)
\end{equation}
where $\delta=1$ is a small margin.

We encourage each thought to focus on a contiguous region in segment space using a total-variation penalty:
\begin{equation}
\mathcal{L}_{cont}
= \frac1{BK}\sum_{b=1}^B\sum_{k=1}^K\sum_{j=2}^M |A_{k,j}^{(b)}-A_{k,j-1}^{(b)}|
\end{equation}

\paragraph{Overall objective.}
Our final training loss is:
\begin{equation}
\mathcal{L}_{total}
= \mathcal{L}_{ce}
+ \lambda_{mono}\mathcal{L}_{mono}
+ \lambda_{cont}\mathcal{L}_{cont}
\end{equation}
where $\lambda_{mono} = 0.1$ and $\lambda_{cont} = 0.2$ control the strength of structural regularization. All components (encoder, thinking module, and decoder) are trained end-to-end with sentence-level supervision. 

\section{LC-HKSLT Dataset}
Despite recent progress on gloss-free SLT, existing benchmarks are often limited in scale or do not fully reflect realistic deployment, where only videos and sentence-level translations (without glosses or SLR vocabularies) are available and may be noisy. This setting places greater emphasis on robust cross-modal reasoning and grounding. To study large-scale, gloss-free SLT in a realistic setting, we construct \textbf{LC-HKSLT}, a Hong Kong Sign Language corpus curated from broadcast-style public briefings with continuously visible interpreters. It contains 1,311 hours (432K clips) and provides only sentence-level supervision (no glosses or SLR vocabularies), matching our intended deployment regime. Full details are provided in Appendix~\ref{app:lchkslt}.

\begin{table*}[t!]
  \centering
    \small
    \setlength{\tabcolsep}{5pt}
    \resizebox{1\linewidth}{!}{
    \begin{tabular}{l|cc|ccccc|ccccc}
       
    \toprule
    { \multirow{2}{*}{\textbf{Method}}} & \multicolumn{2}{c|}{\textbf{Modality}} & \multicolumn{5}{c|}{\textbf{How2Sign}}& \multicolumn{5}{c}{\textbf{OpenASL}}\\
    \cmidrule(lr){2-3}  \cmidrule(lr){4-8} \cmidrule(lr){9-13} 
    & Pose & RGB & B@1  & B@2 & B@3 & B@4 & ROUGE & B@1  & B@2 & B@3 & B@4 & ROUGE \\
    \midrule

    GloFE-VN~\cite{lin2023gloss} &  \Pmodal  & 14.94 & 7.27 & 3.93 & 2.24 & 12.61 & 21.56 & 12.74 & 9.05 & 7.06 & 21.75\\
    T5-SLT~\cite{uthus2023youtubeasl} &   \Pmodal & 14.96 & 5.11 & 2.26 & 1.22 & - & - & - & - & - & - \\
    \color{gray}{T5-SLT-YT}~\cite{uthus2023youtubeasl} & \Pmodal & 37.82 & 24.13 & 16.92 & 12.39 & - & - & - & - & - & -  \\
    {VAP}~\citep{jiao2024visual}  & \Pmodal  & \textbf{39.22} & - & - & \underline{12.87} & \underline{27.77} & \textbf{45.92} & - & - & \textbf{21.23} & \textbf{41.38}\\
    I3D-transformer~\cite{shi2022open} & \Rmodal & - & - & - & - & - & 18.31 & 10.15 & 7.19 & 5.66 & 18.64 \\
    OpenASL~\cite{shi2022open}  & \Rmodal & - & - & - & - & - & 20.92 & 12.08 & 8.59 & 6.72 & 21.02\\
    TF-H2S~\cite{alvarezsign}&  \Rmodal  &17.40 & 7.69 & 3.97 & 2.21 &- & - & - & - & - & - \\
    SLT-IV~\cite{tarres2023sign} & \Rmodal  &34.01 & 19.30 & 12.18 & 8.03& - & - & - & - & - & - \\

    {C$^2$RL}~\citep{chen2024c2rl}  & \Rmodal  & 29.07 & 18.56 & 12.92 & 9.37 & 27.02 &  31.46 & 21.85 & 16.58 & 13.21 &31.36\\

    \midrule
    
    \textbf{\modelname~(Ours)} &  \Rmodal& \underline{36.65} & \textbf{20.96} & \textbf{16.27} & \textbf{13.39} & \textbf{27.85}  & \underline{38.10} & \textbf{26.33} & \textbf{23.12} & \underline{19.55} & \underline{38.78}\\
    \bottomrule
    \end{tabular}}
    \vspace{-0.7em}
    \caption{How2Sign and OpenASL datasets. {\color{gray}{T5-SLT-YT}} denotes training on YouTube-ASL and then fine-tuning on the 30-hour split How2Sign. The best results are highlighted as \textbf{bold}, and the second-best are \underline{underlined}.}
    \vspace{-1.0em}
    \label{tab:sota2}
\end{table*}

\section{Experiments}
\subsection{Experimental Setup}
\label{sec:impl}

\textbf{Datasets.} Following prior works~\cite{chen2022simple, zhou2023gloss, zhou2024scaling}, we conduct experiments on five SLT benchmarks: PHOENIX-2014T (DGS)~\cite{camgoz2018neural}, CSL-Daily (CSL)~\cite{zhou2021improving}, How2Sign (ASL)~\cite{duarte2021how2signlargescalemultimodaldataset}, OpenASL (ASL)~\cite{shi2022openasl}, and our LC-HKSLT corpus.



\noindent\textbf{Evaluation Metrics.} In line with previous studies~\cite{chen2022simple, chen2022two, zhou2023gloss}, we report BLEU~\cite{papineni2002bleu} and ROUGE‑L~\cite{lin2004rouge} for SLT. BLEU‑$n$ measures $n$‑gram precision; we present BLEU‑1 to BLEU‑4 (B1-B4). ROUGE‑L computes the F1 score based on the longest common subsequence between the hypothesis and the reference.


\begin{table}[t!]
  \centering
  \setlength{\tabcolsep}{10pt}
  \resizebox{1\linewidth}{!}{
  \begin{tabular}{@{}l|ccccc@{}}
    \toprule
    { Method}    &  B@1 &  B@2 &  B@3 &  B@4 & ROUGE \\
    \midrule
     NSLT~\citep{camgoz2018neural} &29.91   & 19.56& 13.02 & 8.65 & 9.67 \\
     TSPNet~\cite{li2020tspnet} & 31.36 &  23.19 & 17.27 & 12.83  &   38.37\\
    GASLT~\citep{yin2023gloss} & \underline{38.65} & 28.27  &\underline{24.05} & 19.40 & 44.92\\
    GFSLT-VLP~\citep{zhou2023gloss} & 37.37 &  \underline{29.73} & 23.03 &   \underline{19.64} & \underline{45.05}\\
      \midrule
      \textbf{\modelname~(Ours)}  & \textbf{39.12} & \textbf{31.75}&  \textbf{25.93}& \textbf{21.15} & \textbf{47.87}\\
      \textbf{\modelname~(Ours)$^{\dagger}$}  & 45.33 & 38.01 &  31.83 & 30.22 & 60.01 \\
     \bottomrule
  \end{tabular}}
  \vspace{-0.7em}
  \caption{SLT results on LC-HKSLT. We report scores for four publicly available open-source gloss-free baselines and our \modelname. $\dagger$ denotes a variant that is pre-trained on the remaining LC-HKSLT data and then fine-tuned on the 30-hour split.}
  \label{tab:hkslt}
  \vspace{-1.85em}
\end{table}

\subsection{Comparison with SOTA Methods}
Tab.~\ref{tab:sota} and~\ref{tab:sota2} compare \modelname\ with representative gloss-free SLT methods across five benchmarks. \modelname\ achieves the best gloss-free BLEU-4 on all datasets and attains the highest ROUGE on PHOENIX14T, How2Sign, OpenASL, and LC-HKSLT, while remaining competitive on CSL-Daily. On PHOENIX14T, \modelname\ reaches \textbf{27.22} BLEU-4 and 54.50 ROUGE, surpassing the strongest prior gloss-free method C$^2$RL (26.75 / 50.96) as well as recent LLM-assisted baselines. On CSL-Daily, it achieves \textbf{23.92} BLEU-4 and 50.99 ROUGE, improving over C$^2$RL and slightly outperforming VAP in BLEU-4. Notably, larger gains are observed on large-scale datasets, with BLEU-4 improving from 9.37 to \textbf{13.39} on How2Sign and from 13.21 to \textbf{19.55} on OpenASL, accompanied by consistent ROUGE improvements. On LC-HKSLT, \modelname\ establishes a new state of the art among publicly available methods, achieving \textbf{21.15} BLEU-4 and \textbf{47.87} ROUGE.
Moreover, due to the limited scale of existing public SLT benchmarks, most prior methods cannot perform additional in-domain pre-training under comparable settings. By pre-training \modelname\ on the remaining LC-HKSLT data and fine-tuning on the curated 30-hour split, we obtain substantial additional gains (denoted by $\dagger$), highlighting the critical role of scaling in-domain sign–text data.

\subsection{Ablation Study}

\begin{table}[t]
\centering
\scriptsize
\setlength{\tabcolsep}{5pt}
\begin{tabular}{c ccccccc cc}
\toprule
ID & Thk & Cau & Rtg & DDec & Pr & $\mathcal{L}_m$ & $\mathcal{L}_c$ & B4$\uparrow$ & R$\uparrow$ \\
\midrule
0  & \checkmark & \checkmark & \checkmark & \checkmark & \checkmark & \checkmark & \checkmark & \textbf{27.49} & \textbf{55.90} \\
1  & \xmark     & --         & --         & \checkmark & \xmark     & \xmark     & \xmark     & 25.30 & 51.20 \\
2  & \checkmark & \xmark     & \checkmark & \checkmark & \checkmark & \checkmark & \checkmark & 26.50 & 53.60 \\
3  & \checkmark & \checkmark & \xmark     & \checkmark & \checkmark & \checkmark & \checkmark & 26.10 & 53.00 \\
4  & \checkmark & \checkmark & \checkmark & \xmark     & \xmark     & \checkmark & \checkmark & 26.20 & 53.10 \\
5  & \checkmark & \checkmark & \checkmark & \checkmark & \xmark     & \checkmark & \checkmark & 26.60 & 53.90 \\
6  & \checkmark & \checkmark & \checkmark & \checkmark & \checkmark & \xmark     & \checkmark & 26.70 & 54.00 \\
7  & \checkmark & \checkmark & \checkmark & \checkmark & \checkmark & \checkmark & \xmark     & 26.75 & 54.05 \\
8  & \checkmark & \checkmark & \checkmark & \checkmark & \checkmark & \xmark     & \xmark     & 26.20 & 53.20 \\
\bottomrule
\end{tabular}
\caption{Key element ablation in SignThought. Thk: latent thinking module; Cau: causal self-attn on thoughts; Rtg: structured routing; DDec: dual-stream decoder; Pr: prior injection; $\mathcal{L}_m/\mathcal{L}_c$: monotonicity/contiguity regularizers. ``--'' denotes not applicable.}
\label{tab:ablation_key}
\vspace{-2.5em}
\end{table}

\paragraph{Impact of Key Components in \modelname} We conduct ablation studies on the PHOENIX14T dev set (Tab.~\ref{tab:ablation_key}); additional results are reported in the Appendix. Removing the latent thinking module (ID~1) leads to the largest performance drop, confirming the necessity of an explicit intermediate reasoning interface. Disabling causal thought updates (ID~2) further degrades accuracy, indicating that chain-structured reasoning is superior to unordered latent representations. Replacing structured routing with soft routing (ID~3) and simplifying the dual-stream decoder to a single stream (ID~4) both impair performance, highlighting the importance of explicit evidence allocation and plan-then-ground decoding. Removing thought-guided prior injection (ID~5) results in a smaller yet consistent decline, suggesting its role in emphasizing relevant temporal evidence. Finally, ablations of $\mathcal{L}{mono}$ and $\mathcal{L}{cont}$ (IDs~6--8) yield limited individual impact but a pronounced drop when jointly removed, implying complementary effects in stabilizing coherent thought-to-evidence alignment.


\begin{table}[t]
\centering
\scriptsize
\setlength{\tabcolsep}{5pt}
\renewcommand{\arraystretch}{1}
\begin{tabular}{@{}lcc|lcc@{}}
\toprule
\multicolumn{3}{c|}{\shortstack[c]{\textbf{Vary $K$} fix $L{=}2$, $M{=}16$}} &
\multicolumn{3}{c}{\shortstack[c]{\textbf{Vary $L$} fix $K{=}8$, $M{=}16$}} \\
\cmidrule(r){1-3}\cmidrule(l){4-6}
$K$ & B4$\uparrow$ & R$\uparrow$ & $L$ & B4$\uparrow$ & R$\uparrow$ \\
\midrule
2  & 25.90 & 52.00 & 1 & 26.70 & 53.90 \\
4  & 26.60 & 53.10 & 2 & \textbf{27.49} & \textbf{55.90} \\
6  & 27.00 & 54.10 & 3 & 27.18 & 54.40 \\
8  & \textbf{27.49} & \textbf{55.90} & 4 & 27.05 & 54.10 \\
10 & 27.10 & 54.30 & 5 & 26.73 & 54.02 \\
\midrule
\multicolumn{6}{c}{\shortstack[c]{\textbf{Vary $M$} fix $K{=}8$, $L{=}2$}} \\
\midrule
\multicolumn{2}{@{}l}{\scriptsize $M$} &
\multicolumn{2}{c}{\scriptsize B4$\uparrow$} &
\multicolumn{2}{c@{}}{\scriptsize R$\uparrow$} \\
\midrule
\multicolumn{2}{@{}l}{4}  & \multicolumn{2}{c}{26.50} & \multicolumn{2}{c@{}}{53.30} \\
\multicolumn{2}{@{}l}{8}  & \multicolumn{2}{c}{26.95} & \multicolumn{2}{c@{}}{54.00} \\
\multicolumn{2}{@{}l}{16} & \multicolumn{2}{c}{\textbf{27.49}} & \multicolumn{2}{c@{}}{\textbf{55.90}} \\
\multicolumn{2}{@{}l}{32} & \multicolumn{2}{c}{27.12} & \multicolumn{2}{c@{}}{54.30} \\
\multicolumn{2}{@{}l}{64} & \multicolumn{2}{c}{26.80} & \multicolumn{2}{c@{}}{53.70} \\
\bottomrule
\end{tabular}
\caption{Hyperparameter/structure-strength ablations.}
\label{tab:ablation_strength}
\vspace{-2.5em}
\end{table}

\paragraph{Hyperparameter and Structure-Strength Ablations.} On PHOENIX14T, we vary one factor at a time while keeping the rest of the full model unchanged. We sweep the number of thoughts $K\!\in\!\{2,4,6,8,10\}$ (reasoning slots), the thinking depth $L\!\in\!\{1,2,3,4,5\}$ (refinement steps), and the number of segment tokens $M\!\in\!\{4,8,16,32,64\}$ (evidence granularity for routing) to study the trade-off between reasoning capacity and routing complexity and to locate a practical sweet spot.

\begin{table}[htbp]
\centering
\tiny
\setlength{\tabcolsep}{2pt}
\renewcommand{\arraystretch}{1.15}
\begin{tabularx}{\columnwidth}{l X}
\hline
\textbf{Reference}: &
\begin{CJK}{UTF8}{gbsn}我完全看不出这两张照片有什么差别。\end{CJK}
\,(I totally can’t tell what the difference is between these two photos.)\\

\textbf{GFSLT-VLP}: &
\begin{CJK}{UTF8}{gbsn}\wrong{我觉得这}\corr{照片有}\wrong{问题}。\end{CJK}
\,(\wrong{I think there’s something wrong with} \corr{this photo}.)\\

\textbf{\modelname}: &
\begin{CJK}{UTF8}{gbsn}\para{我真的}\corr{看不出这两张照片有什么}\para{区别}。\end{CJK}
\,(\para{I really can’t} \corr{tell what the} \para{distinction between} \corr{these two photos}.)\\

\hline
\textbf{Reference}: &
\begin{CJK}{UTF8}{gbsn}这场大雨已经下了一整天了。\end{CJK}
\,(This heavy rain has been falling all day.)\\

\textbf{GFSLT-VLP}: &
\begin{CJK}{UTF8}{gbsn}\wrong{外面正在}\para{下雨}\wrong{一天}。\end{CJK}
\,(\wrong{It’s} \para{raining} \wrong{outside for one day}.)\\

\textbf{\modelname}: &
\begin{CJK}{UTF8}{gbsn}\corr{这场大雨}\para{下了整整一天}。\end{CJK}
\,(\corr{This heavy rain has been falling for} \para{the whole day}.)\\
\hline

\textbf{Reference}: &
\begin{CJK}{UTF8}{gbsn}晚上我请你去吃肯德基。\end{CJK}
\,(I’ll treat you to KFC tonight.)\\

\textbf{GFSLT-VLP}: &
\begin{CJK}{UTF8}{gbsn}\corr{晚上}\wrong{一起去吃饭吧}。\end{CJK}
\,(\wrong{Let’s go out for dinner} \corr{tonight}.)\\

\textbf{\modelname}: &
\begin{CJK}{UTF8}{gbsn}\para{今晚}\corr{我请你去吃肯德基}。\end{CJK}
\,(\para{Tonight}, \corr{I’ll treat you to KFC.})\\

\hline
\textbf{Reference}: &
\begin{CJK}{UTF8}{gbsn}走在一对情侣旁边，我感觉自己很多余。\end{CJK}
\,(Walking next to a couple, I feel like a third wheel.)\\

\textbf{GFSLT-VLP}: &
\begin{CJK}{UTF8}{gbsn}\para{我走}\wrong{在他们之间}，\corr{感觉}\wrong{不太自在}。\end{CJK}
\,(\corr{Walking} \wrong{between them}, \wrong{I don’t} \corr{feel} \wrong{very comfortable}.)\\

\textbf{\modelname}: &
\begin{CJK}{UTF8}{gbsn}\corr{走在}\para{那对情侣旁边}，\corr{我觉得自己特别多余}。\end{CJK}
\,(\para{Walking next to that couple}, \corr{I feel like such a third wheel}.)\\

\hline

\end{tabularx}
\caption{Translation results. Correct answers are marked in green, semantically similar but rephrased answers are displayed in blue, and incorrect answers are indicated in red.}
\label{tbl:qualitative}
\vspace{-2.85em}
\end{table}

\subsection{Qualitative Results}
Table~\ref{tbl:qualitative} presents representative CSL-Daily cases comparing \modelname\ with GFSLT-VLP, a only recent open-source baseline for gloss-free SLT. Following the color scheme in the caption, \modelname\ consistently preserves the core semantics of the reference while allowing minor paraphrases (blue), whereas GFSLT-VLP more often exhibits semantic drift or missing key information (red), e.g., over-generalizing ``difference between two photos'' to a vague ``something wrong,'' weakening duration expressions such as ``all day,'' or dropping salient entities (e.g., \textit{KFC}). These examples qualitatively support our claim that the proposed latent reasoning and evidence allocation help maintain faithful, grounded translations, especially for sentences requiring fine-grained semantic composition.

\section{Conclusion}
We propose \modelname, a reasoning-driven framework for gloss-free sign language translation that introduces Cross-Modal Latent Thoughts and a Latent Chain-of-Thought to bridge continuous sign videos and text generation without gloss supervision. \modelname\ achieves SOTA performance across five benchmarks, demonstrating that latent-thought reasoning provides a scalable alternative to gloss annotations. We further introduce LC-HKSLT, a large-scale Cantonese SLT dataset to facilitate evaluation in realistic settings. Future work will explore stronger reasoning supervision, improved training and inference efficiency, and extensions to broader sign languages and open-world scenarios.
\section*{Limitation}
Although SignThought introduces an ordered thought chain, the ``thinking'' process in our framework remains latent rather than explicit. The intermediate thoughts are continuous hidden states that are only indirectly learned from the final translation objective, rather than being verbalized, externally supervised, or exposed as human-interpretable reasoning steps. As a result, while the model benefits from improved planning and grounding, its intermediate reasoning is still difficult to directly inspect, verify, or control. In particular, we cannot guarantee that each latent thought corresponds to a stable semantic concept or a human-recognizable reasoning unit, and error analysis remains largely outcome-based at the final translation level. Therefore, the current framework should be viewed as a step toward reasoning-aware sign language translation, rather than a system that already produces explicit and fully interpretable reasoning traces. An important direction for future work is to bridge latent planning with more explicit forms of reasoning, such as textual rationales, gloss-like abstractions, or controllable semantic plans~\cite{dong2026neureasonerexplainablecontrollableunified}.

\section*{Acknowledgement}
The research described in this paper was supported by the National Natural Science Foundation of China (Grant No. 62372314). This work was also supported by computational resources provided by The Centre for Large AI Models (CLAIM) of The Hong Kong Polytechnic University.


\bibliography{custom}

\newpage
\appendix
\label{sec:appendix}
\section{Method and Implementation Details}
\subsection{Implementation Details.}
\label{appdix.A}
All experiments are conducted on a single NVIDIA RTX A6000 GPU. We implement our model in PyTorch and train all components (sign encoder, Latent CoT thinking module, and dual-stream decoder) end-to-end for 200 epochs. Unless otherwise specified, we use a shared model dimension $d{=}256$ throughout the encoder, thinking module, and decoder. All datasets use the same pre-extracted visual features of dimension 1024, following the feature extraction pipeline used in~\cite{camgoz2020sign,voskou2021stochastic}, which adopts an identical approach with an open-sourced pretrained Inception network. The sign encoder consists of $N_{\text{enc}}=2$ stacked encoder layers, and the Latent CoT thinking module contains $L=2$ thinking layers with $K=8$ learnable thought slots. The decoder is a standard autoregressive Transformer with $N_{dec}=2$ layers, where each layer performs masked self-attention, thinking cross-attention to the thought chain, and visual cross-attention to encoder features (Sec.~\ref{sec:decoder}). 

For the optional thought-guided evidence routing, we reuse the cached routing variables from the thinking module and apply them as a soft bias in the decoder's visual cross-attention.  We initialize linear/embedding weights with Xavier initialization and apply dropout with a rate of 0.1 to attention and feed-forward sublayers. We use layer normalization in all Transformer-style blocks.  We optimize with Adam (lr $=1\times10^{-3}$, $\beta_1{=}0.9$, $\beta_2{=}0.998$) and a batch size of 32. We train the model with weight decay $=3\times 10^{-3}$ and label smoothing $=0.1$. We use a plateau-based learning rate scheduler and apply a warmup phase of 2000 steps before the scheduler takes effect.

We evaluate on the validation set periodically and reduce the learning rate by a factor of 0.8 if the validation metric does not improve for several consecutive checks, stopping when the learning rate drops below $1\times10^{-4}$. At test time, we use beam search for translation decoding and tune decoding hyperparameters on the development set. We sweep the beam size $b \in [1,10]$ and length-penalty coefficient $a \in \{-1,0,1,2,3,4,5\}$ on the dev set; the best $(b,a)$ is used for test-time decoding.

\subsection{Soft Segmentation via Soft Boundaries}
\label{app:soft_seg}

Given encoder frame features $\mathbf{E}=\{e_t\}_{t=1}^{T_s}$, we construct $M$ segment tokens
$\mathbf{S}=\{S_j\}_{j=1}^{M}$ by predicting \emph{soft} temporal windows with learnable boundaries
(per instance). Let $\bar m_t\in\{0,1\}$ denote a multiplicative padding mask over source frames
($\bar m_t=0$ for padded frames and $\bar m_t=1$ otherwise). Define the valid length
$L_{\text{vaild}}=\sum_{t=1}^{T_s}\bar m_t$ (we assume $L_{\text{vaild}}\ge 1$).

We first obtain a clip-level summary by masked mean pooling:
\begin{equation}
z=\frac{\sum_{t=1}^{T_s}\bar m_t\, e_t}{L_{\text{vaild}}+\epsilon_{\mathrm{num}}}\in\mathbb{R}^{d}.
\end{equation}
We then predict positive segment lengths $\rho\in\mathbb{R}^{M}$ with a boundary MLP:
\begin{align}
\rho=\mathrm{softplus}(\mathrm{MLP}_{\mathrm{bd}}(z)), 
\\
\pi=\frac{\rho}{\sum_{j=1}^{M}\rho_j+\epsilon_{\mathrm{num}}},
\end{align}
and convert proportions into cumulative boundaries in continuous time over valid frames:
\begin{equation}
\tau_0=1,\ \ 
\tau_j = 1+(L_{\text{vaild}}-1)\sum_{i=1}^{j}\pi_i,\quad j=1,\ldots,M,
\end{equation}
so $\tau_M\approx L_{\text{vaild}}$ and each instance can have different segment durations even when $M$ is fixed.

\paragraph{Soft window membership and weights.}
Let $\hat t=\sum_{u=1}^{t}\bar m_u$ denote the valid-frame rank of position $t$.
For each segment $j$, we define a soft window membership over discrete frames:
\begin{equation}
u_{j,t}=\sigma(\gamma(\hat t-\tau_{j-1}))-\sigma(\gamma(\hat t-\tau_j)),\ \  t=1,\ldots,T_s,
\end{equation}
where $\sigma(\cdot)$ is the sigmoid and $\gamma>0$ controls boundary softness (we set $\gamma=1$).
We apply the padding mask and normalize over frames:
\begin{equation}
u_{j,t}\leftarrow \bar m_t\,u_{j,t},\ \ 
(\mathbf W_{seg})_{j,t}=\frac{u_{j,t}}{\sum_{t'=1}^{T_s}u_{j,t'}+\epsilon_{\mathrm{num}}},
\end{equation}
so each row of $\mathbf W_{seg}\in\mathbb{R}^{M\times T_s}$ sums to $1$.

\paragraph{Segment tokens.}
Finally, segment tokens are obtained by weighted pooling:
\begin{equation}
S_j=\sum_{t=1}^{T_s}(\mathbf W_{seg})_{j,t}\,e_t,\qquad j=1,\ldots,M.
\end{equation}
We use $\epsilon_{\mathrm{num}}=10^{-6}$ throughout for numerical stability.

\subsection{Injecting routed summaries $p_k$ into RoutedXAttn}
\label{app:pk_injection}

We implement conditioning on $p_k$ by converting it into a routing-dependent bias over source time
steps and adding it to the cross-attention logits. For each head $h\in\{1,\dots,H\}$ with head
dimension $d_h=d/H$, we compute standard logits:
\begin{equation}
\ell^{(h)}_{k,t}=\frac{\left(W_Q^{(h)}\tilde c_k\right)^\top\left(W_K^{(h)}e_t\right)}{\sqrt{d_h}},
\end{equation}
and add a $p_k$-dependent bias before softmax:
\begin{equation}
\tilde \ell^{(h)}_{k,t}= \ell^{(h)}_{k,t} + \lambda_p\, b^{(h)}_{k,t}(p_k,\mathbf{E}),
\label{eq:pk_bias_logits}
\end{equation}
where $\lambda_p\ge 0$ controls the bias strength (we set $\lambda_p=1$ in all experiments).

\paragraph{Attention with hard padding mask.}
Let $m\in\{0,-\infty\}^{T_s}$ be the additive attention mask over the $T_s$ source frames, where for each $t\in\{1,\ldots,T_s\}$, $m_t=0$ if frame $t$ is valid (non-padded) and $m_t=-\infty$ otherwise; then:
\begin{align}
a^{(h)}_{k,t} &= \mathrm{softmax}_{t}\!\left(\tilde \ell^{(h)}_{k,t} + m_{t}\right),\\
o^{(h)}_k &= \sum_{t=1}^{T_s}a^{(h)}_{k,t}\left(W_V^{(h)}e_t\right).
\end{align}

\paragraph{A minimal content-based bias.}
A simple differentiable instantiation is:
\begin{equation}
b^{(h)}_{k,t}(p_k,\mathbf{E})
=\frac{\left(W_P^{(h)}p_k\right)^\top\left(W_B^{(h)}e_t\right)}{\sqrt{d_h}}.
\label{eq:pk_bias_content}
\end{equation}

\paragraph{Optional routing-derived temporal prior.}
If segment-to-frame weights $\mathbf{W}_{seg}\in\mathbb{R}^{M\times T_s}$ are available (row-normalized
over frames), we can derive a purely routing-based temporal prior:
\begin{align}
\mathbf r_k=\mathbf A_{k,:}\mathbf W_{seg}\in\mathbb{R}^{1\times T_s},
\\
r_{k,t}=\sum_{j=1}^{M}A_{k,j}(\mathbf W_{seg})_{j,t}.   
\end{align}
We then inject it as an additive log-prior:
\begin{equation}
b^{(h)}_{k,t}\;\leftarrow\; b^{(h)}_{k,t} + \log(r_{k,t}+\epsilon_{\mathrm{num}}),
\label{eq:pk_bias_routing}
\end{equation}
with $\epsilon_{\mathrm{num}}=10^{-6}$.

\subsection{Injecting the temporal prior $\mathbf{w}$ into grounding cross-attention}
\label{app:w_injection}

We compute a token-to-frame prior $\mathbf{w}\in\mathbb{R}^{T_t\times T_s}$ as:
\begin{equation}
\boldsymbol{\beta} = \boldsymbol{\alpha}\mathbf{A},\qquad
\mathbf{w} = \boldsymbol{\beta}\mathbf{W}_{seg},
\end{equation}
where $\boldsymbol{\alpha}\in\mathbb{R}^{T_t\times K}$ are head-averaged decoder-to-thought attention
weights (row-normalized over $K$ thoughts). Under non-negativity and approximate row normalization of
$\boldsymbol{\alpha},\mathbf{A},\mathbf{W}_{seg}$, each row $\mathbf{w}_{t,\cdot}$ is an (approximately)
normalized soft prior over source frames.

For grounding cross-attention at layer $l$, logits are:
\begin{equation}
\ell^{(h)}_{t,s}=\frac{\left(W_Q^{(h)}h^{(l)}_{\mathrm{think},t}\right)^\top\left(W_K^{(h)}e_s\right)}{\sqrt{d_h}}.
\end{equation}
When enabled, we add a soft bias:
\begin{equation}
\tilde \ell^{(h)}_{t,s}= \ell^{(h)}_{t,s} + \lambda_w \log(\mathbf{w}_{t,s}+\epsilon_{\mathrm{num}}),
\end{equation}
where $\lambda_w\ge 0$ (we set $\lambda_w=1$ in all experiments). We then apply the standard additive
source mask $m_s\in\{0,-\infty\}$:
\begin{equation}
a^{(h)}_{t,s}=\mathrm{softmax}_{s}\!\left(\tilde \ell^{(h)}_{t,s}+m_s\right).
\end{equation}

\label{app:pk_injection}

Let the $K$ thought tokens at layer $\ell$ be $\tilde{\mathbf{C}}^{(\ell)}=[\tilde c_1,\dots,\tilde c_K]^\top \in \mathbb{R}^{K\times d}$.
Let the encoder (frame-level) features be $\mathbf{E}=[e_1,\dots,e_{T_s}]^\top \in \mathbb{R}^{T_s\times d}$.
Let the segment tokens be $\mathbf{S}=[S_1,\dots,S_M]^\top \in \mathbb{R}^{M\times d}$, and the Sinkhorn routing matrix be
$\mathbf{A}\in\mathbb{R}^{K\times M}$ with non-negative entries and approximately row-normalized
($\sum_{j=1}^{M}A_{k,j}\approx 1$).
The routed summary for thought $k$ is
\begin{equation}
p_k=\sum_{j=1}^{M}A_{k,j}S_j,\qquad p_k\in\mathbb{R}^{d},
\end{equation}
and we denote $\mathbf p=\{p_k\}_{k=1}^{K}$.

Conditioned on $p_k$, $\mathrm{RoutedXAttn}$ performs cross-attention from the thought tokens to the encoder features, with a $p_k$-dependent bias that favors frames consistent with the routed segment content (or equivalently, the routing induced by $\mathbf{A}$).

Concretely, for each head $h\in\{1,\dots,H\}$ with head dimension $d_h=d/H$, we compute standard cross-attention logits
\begin{equation}
\ell^{(h)}_{k,t}=\frac{\left(W_Q^{(h)}\tilde c_k\right)^\top\left(W_K^{(h)}e_t\right)}{\sqrt{d_h}},
\end{equation}
and add a $p_k$-dependent bias term $b^{(h)}_{k,t}(p_k,\mathbf{E})$ before the softmax:
\begin{equation}
\tilde \ell^{(h)}_{k,t}= \ell^{(h)}_{k,t} + \lambda_p\, b^{(h)}_{k,t}(p_k,\mathbf{E}).
\label{eq:pk_bias_logits_}
\end{equation}
where $\lambda_p\ge 0$ controls the strength of the routed-summary bias (we set $\lambda_p=1$ in all experiments unless otherwise specified).
The attention weights and outputs follow the standard definition:
\begin{align}
a^{(h)}_{k,t} &= \mathrm{softmax}_{t}\!\left(\tilde \ell^{(h)}_{k,t} + m_{t}\right),\\
o^{(h)}_k &= \sum_{t=1}^{T_s}a^{(h)}_{k,t}\left(W_V^{(h)}e_t\right),
\end{align}
where $\mathrm{softmax}_{t}$ denotes normalization over the source time index $t$.
Here $m_t=-\infty$ for masked (padded) time steps and $m_t=0$ otherwise; thus the mask remains a hard constraint.

A minimal instantiation is a content-based compatibility between the routed summary and each source frame:
\begin{equation}
b^{(h)}_{k,t}(p_k,\mathbf{E})
=\frac{\left(W_P^{(h)}p_k\right)^\top\left(W_B^{(h)}e_t\right)}{\sqrt{d_h}},
\label{eq:pk_bias_content}
\end{equation}
which increases attention to frames whose features align with the routed segment summary.

Alternatively, when the segmentation weights $\mathbf{W}_{seg}\in\mathbb{R}^{M\times T_s}$ are available (row-normalized over frames), we can additionally derive a purely routing-based temporal prior
\begin{equation}
r_{k,t}=\sum_{j=1}^{M}A_{k,j}\,(\mathbf W_{seg})_{j,t},\qquad t=1,\ldots,T_s,
\end{equation}
We denote the resulting temporal prior vector by $\mathbf r_k\in\mathbb{R}^{T_s}$ with entries $(\mathbf r_k)_t=r_{k,t}$.
and inject it as a log-prior bias:
\begin{equation}
b^{(h)}_{k,t}\;\leftarrow\; b^{(h)}_{k,t} + \log(r_{k,t}+\epsilon_{\mathrm{num}}).
\label{eq:pk_bias_routing}
\end{equation}
This realizes the ``equivalently, the routing induced by $\mathbf{A}$'' view: both $p_k$ and $\mathbf r_k$ are functions of the same routing variables,
with $p_k$ summarizing \emph{what} evidence is assigned, and $\mathbf r_k$ summarizing \emph{where in time} it is expected to lie.

We write Eqs.~\eqref{eq:pk_bias_logits}--\eqref{eq:pk_bias_routing} as
\begin{equation}
\hat{\mathbf{C}}^{(\ell)} = \mathrm{RoutedXAttn}\!\left(\tilde{\mathbf{C}}^{(\ell)}, \mathbf{E}, \mathbf{p}\right),
\end{equation}
where $\mathrm{RoutedXAttn}$ performs standard multi-head cross-attention from thoughts to encoder features with the above biasing mechanism.

\subsection{Injecting the temporal prior \texorpdfstring{$\mathbf{w}$}{w} into the Dual-Stream Decoder grounding cross-attention}
\label{app:w_injection}

Let $T_t$ be the target length. We compute a token-to-frame prior $\mathbf{w}\in\mathbb{R}^{T_t\times T_s}$ per instance as
\begin{equation}
\boldsymbol{\beta} = \boldsymbol{\alpha}\mathbf{A},\qquad
\mathbf{w} = \boldsymbol{\beta}\mathbf{W}_{seg},
\label{eq:w_def_app}
\end{equation}
where $\boldsymbol{\alpha}\in\mathbb{R}^{T_t\times K}$ is the head-averaged \emph{attention weights} (row-normalized over $K$ thoughts) from the decoder token positions to the $K$ thought tokens.
With non-negative $\boldsymbol{\alpha},\mathbf{A},\mathbf{W}_{seg}$ and approximate normalization (rows summing to $\approx 1$), each row vector $\mathbf w_t\in\mathbb{R}^{T_s}$ (the $t$-th row of $\mathbf w$) forms an (approximately) normalized soft temporal prior over source time steps.

In the dual-stream decoder, we compute $\boldsymbol{\alpha}$ from the thought-attention sublayer first (standard attention producing weights), then form $\mathbf{w}$ via Eq.~\eqref{eq:w_def_app}, and finally use $\mathbf{w}$ in the grounding (encoder) cross-attention sublayer as a bias. This ensures $\mathbf{w}$ is available before computing the grounded attention distribution.

Let $\mathbf{H}^{(l)}_{\mathrm{think}}\in\mathbb{R}^{T_t\times d}$ denote the decoder states entering the grounding cross-attention at layer $l$.
Vanilla cross-attention to encoder features uses logits
\begin{equation}
\ell^{(h)}_{t,s}=\frac{\left(W_Q^{(h)}h^{(l)}_{\mathrm{think},t}\right)^\top\left(W_K^{(h)}e_s\right)}{\sqrt{d_h}}.
\end{equation}
When the routing prior is enabled, we add a soft bias derived from $\mathbf{w}$ before the softmax:
\begin{equation}
\tilde \ell^{(h)}_{t,s}= \ell^{(h)}_{t,s} + \lambda_w \log(\mathbf{w}_{t,s}+\epsilon_{\mathrm{num}}),
\label{eq:w_bias_logits}
\end{equation}
where $\lambda_w\ge 0$ controls the strength of the temporal prior bias (we set $\lambda_w=1$ in all experiments unless otherwise specified). Followed by the standard source mask $m_s$ (padded frames remain excluded):
\begin{align}
a^{(h)}_{t,s} &= \mathrm{softmax}_{s}\!\left(\tilde \ell^{(h)}_{t,s}+m_{s}\right),\\
\mathbf{H}^{(l)}_{\mathrm{vis},t} &= \mathrm{Concat}_h\left(\sum_{s=1}^{T_s}a^{(h)}_{t,s}W_V^{(h)}e_s\right)W_O.
\end{align}
where $\mathrm{softmax}_{s}$ denotes normalization over the source time index $s$.
Thus $\mathbf{w}$ steers the grounding cross-attention toward time regions consistent with the routed thought mixture implied by $\boldsymbol{\alpha}$, while preserving the standard Transformer decoder structure.
Setting $\lambda_w=0$ disables the mechanism without changing the model I/O or the decoder interface.

\section{Additional Ablations}
\label{app:ablation}

\subsection{Ablations on the Evidence Fabric}
\label{app:ablation:fabric}

To verify that the gains come from multi-granular evidence rather than extra computation, we ablate the evidence fabric while keeping the routing and the dual-stream decoder unchanged, and only vary which evidence tokens are provided (Table~\ref{tab:ablation_fabric}). We evaluate: (i) \textbf{Frame-only}, where the fabric consists of frame-level tokens $\mathbf{E}$ only; (ii) \textbf{Segment-only}, where the fabric consists of segment tokens $\mathbf{S}$ only; and (iii) \textbf{Global-only}, where we replace the fabric with a single pooled token $g$ computed by masked mean pooling over $\mathbf{E}$ (used only for this ablation). Frame evidence preserves fine lexical cues but weakens long-range composition; segment evidence supports planning but loses fine details; global-only is the most compressed and thus the weakest.

\begin{table}[htbp]
\centering
\small
\setlength{\tabcolsep}{4pt}
\begin{tabular}{lcc}
\toprule
Fabric Variant (PHOENIX14T) & B4$\uparrow$ & R$\uparrow$ \\
\midrule
Frame-only (no seg) & 26.40 & 53.40 \\
Segment-only (no frame) & 26.00 & 52.90 \\
Global-only (single pooled token) & 24.80 & 50.80 \\
Frame+Segment (Full) & \textbf{27.49} & \textbf{55.90} \\
\bottomrule
\end{tabular}
\caption{Evidence fabric ablations on PHOENIX14T}
\label{tab:ablation_fabric}
\end{table}

\subsection{Additional Analyses: Length Buckets and Interpretability Metrics}
\label{app:analysis:length_interp}

\paragraph{Length-bucket evaluation.}
To further examine whether \modelname\ improves \emph{long-range semantic composition}, we evaluate BLEU-4 under target-length buckets on PHOENIX14T (Table~\ref{tab:length_bucket}).  
We divide the dev set into three equally sized groups by reference length $|\mathbf{y}|$ (in BPE tokens): \textsc{Short} ($|\mathbf{y}| \le 9$), \textsc{Medium} ($10 \le |\mathbf{y}| \le 17$), and \textsc{Long} ($|\mathbf{y}| \ge 18$), corresponding to the lower, middle, and upper tertiles of the dev set length distribution.  

As shown in Table~\ref{tab:length_bucket}, the improvements of \modelname\ over the baseline grow with target length: +0.4 for short sentences, +1.1 for medium, and +2.2 for long sentences.  
This trend suggests that the latent reasoning chain contributes most when translation requires \emph{compositional aggregation of multiple visual evidences} or modeling long-range temporal dependencies, aligning with the intended “plan–then–ground” inductive bias.

\begin{table}[t]
\centering
\small
\setlength{\tabcolsep}{4pt}
\begin{tabular}{lccc}
\toprule
BLEU-4 by $|\mathbf{y}|$ buckets & Short & Medium & Long \\
\midrule
w/o latent thinking & 29.20 & 25.60 & 19.40 \\
\modelname\ (Full) & 29.60 & 26.70 & 21.60 \\
Gain (Full -- Base) & +0.40 & +1.10 & +2.20 \\
\bottomrule
\end{tabular}
\caption{Length-bucket BLEU-4 on PHOENIX14T.}
\label{tab:length_bucket}
\end{table}

\begin{table}[t]
\centering
\small
\setlength{\tabcolsep}{4pt}
\begin{tabular}{lcccc}
\toprule
Method & Entropy$\downarrow$ & MonoViol$\downarrow$ & Span$\downarrow$ & TV$\downarrow$ \\
\midrule
w/o latent thinking & 2.10 & 0.27 & 18.5 & 0.34 \\
\modelname\ (Full) & \textbf{1.45} & \textbf{0.12} & \textbf{14.0} & \textbf{0.22} \\
\bottomrule
\end{tabular}
\caption{Quantitative interpretability metrics on PHOENIX14T.}
\label{tab:interp_metrics}
\end{table}

\paragraph{Quantitative interpretability metrics.}
We additionally quantify whether the latent chain learns selective and ordered thought-to-evidence alignments.  
Metrics are computed from the final-layer thought-to-segment binding matrix $A\in\mathbb{R}^{K\times M}$ on the PHOENIX14T test set and averaged over samples.  
Lower values indicate more structured and interpretable alignments.

\begin{itemize}
\item \textbf{Entropy ($\downarrow$)}: measures selectivity of each thought $k$ as $H_k=-\sum_{j=1}^{M}A_{k,j}\log(A_{k,j}+\epsilon)$, averaged over $k$. Lower entropy indicates sharper attention (less diffuse binding).

\item \textbf{Monotonicity violation (MonoViol, $\downarrow$)}: captures ordering consistency.  
Let $\mu_k=\sum_{j} jA_{k,j}$ be the expected segment index for thought $k$.  
We report the fraction of violations $\text{MonoViol}=\frac{1}{K-1}\sum_{k=1}^{K-1}\mathbb{I}[\mu_{k}>\mu_{k+1}]$,  
where smaller values mean that later thoughts attend to later evidence more consistently.

\item \textbf{Span ($\downarrow$)}: quantifies localization sharpness.  
For each $k$, define coverage indices $j^{(p)}_k$ satisfying $\sum_{j\le j^{(p)}_k}A_{k,j}\ge p$, and compute $\text{Span}_k=j^{(0.95)}_k-j^{(0.05)}_k$.  
Smaller spans imply tighter evidence localization.

\item \textbf{Total variation (TV, $\downarrow$)}: measures fragmentation of attention along time,  
$\text{TV}=\frac{1}{K}\sum_{k=1}^{K}\sum_{j=2}^{M}|A_{k,j}-A_{k,j-1}|$.  
Lower TV indicates smoother, more contiguous alignments.
\end{itemize}

Table~\ref{tab:interp_metrics} reports these metrics, showing consistent improvements across all four aspects.  
\modelname\ exhibits lower entropy (1.45 vs.\ 2.10), fewer ordering violations (0.12 vs.\ 0.27), tighter localization (14.0 vs.\ 18.5), and smoother attention (0.22 vs.\ 0.34).  
Together, these results quantitatively confirm that the latent thought chain promotes coherent, ordered, and interpretable evidence allocation, aligning with the qualitative visualizations in Fig.~2.
\begin{figure*}[htbp]
\centering
    \includegraphics[width=0.85\linewidth]{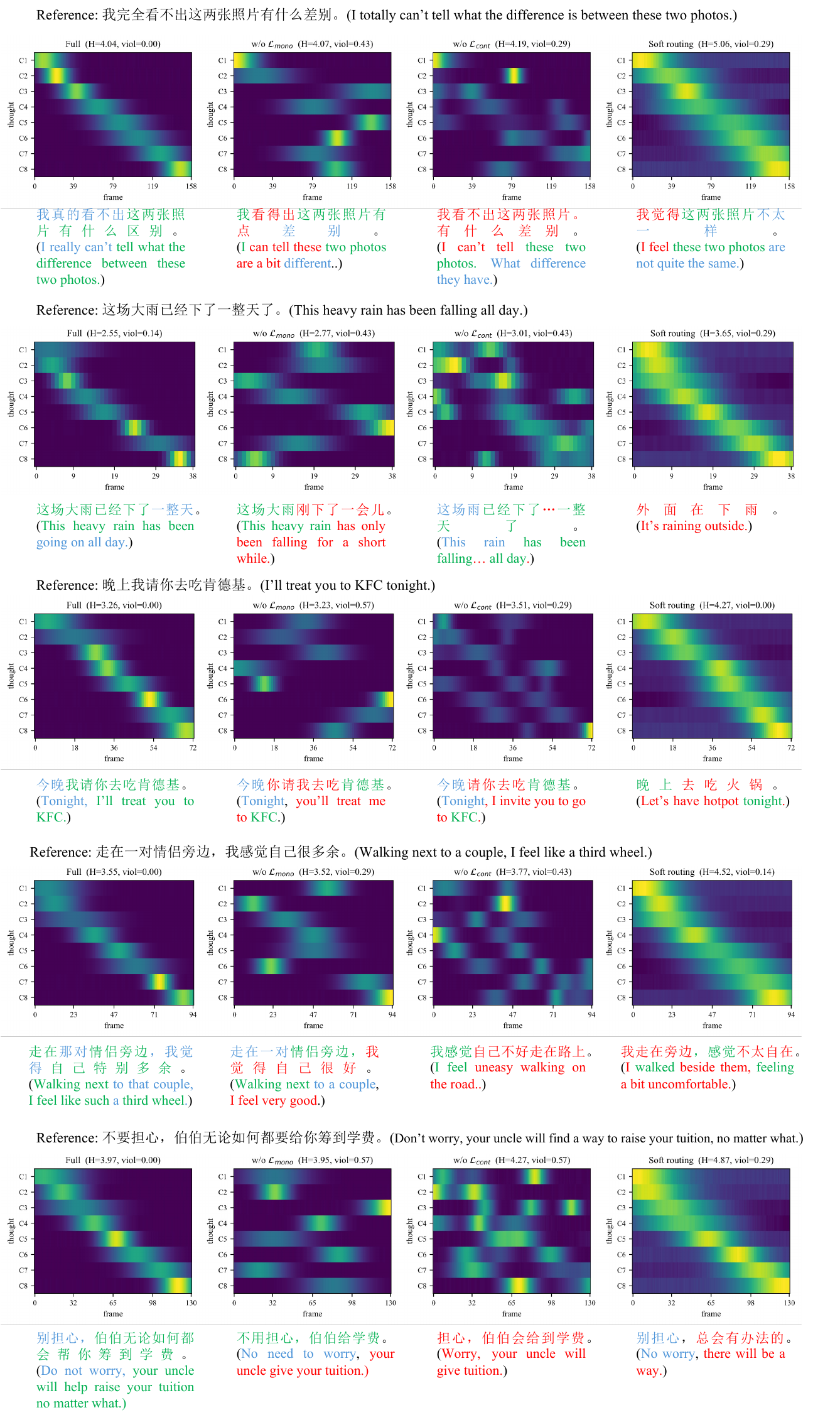}
    \caption{Qualitative visualization on CSL-Daily.}

    \label{fig:qualitative_vis}
\end{figure*}

\subsection{Effect of Structural Regularization Strength}
\label{sec:ablation_mono_strength}

We study the sensitivity of the structured routing regularization to the two loss weights
$\lambda_{\mathrm{mono}}$ and $\lambda_{\mathrm{cont}}$ in
\begin{equation}
\mathcal{L}_{total}
= \mathcal{L}_{ce}
+ \lambda_{\mathrm{mono}}\mathcal{L}_{mono}
+ \lambda_{\mathrm{cont}}\mathcal{L}_{cont}.
\end{equation}
These terms encourage (i) monotonic progression of the thought-to-segment assignment and (ii) reduced fragmentation (contiguity) in segment space, respectively. Concretely, we sweep $\lambda_{\mathrm{mono}} \in \{0, 0.01, 0.05, 0.1, 0.2, 0.5, 1.0\}$ and
$\lambda_{\mathrm{cont}} \in \{0, 0.05, 0.1, 0.2, 0.4\}$ while keeping all other settings fixed.
For clarity, Table~\ref{tab:reg_sweep} reports representative one-dimensional sweeps on the development set:
(i) varying $\lambda_{\mathrm{mono}}$ with $\lambda_{\mathrm{cont}}$ fixed to $0.2$, and
(ii) varying $\lambda_{\mathrm{cont}}$ with $\lambda_{\mathrm{mono}}$ fixed to $0.1$.

Overall, introducing small-to-moderate structural regularization consistently improves translation quality over the no-regularization baseline ($\lambda_{\mathrm{mono}}=\lambda_{\mathrm{cont}}=0$).
Performance peaks at $\lambda_{\mathrm{mono}}=0.1$ with $\lambda_{\mathrm{cont}}=0.2$, while further increasing
either coefficient degrades performance, suggesting over-regularization.
Based on this sweep, we set $\lambda_{\mathrm{mono}}=0.1$ and $\lambda_{\mathrm{cont}}=0.2$ as the default choice
in all experiments.

\begin{table}[htbp]
\centering
\small
\setlength{\tabcolsep}{6pt}
\begin{tabular}{cc|c||cc|c}
\toprule
\multicolumn{3}{c||}{Sweep $\lambda_{\mathrm{mono}}$} &
\multicolumn{3}{c}{Sweep $\lambda_{\mathrm{cont}}$} \\
\midrule
$\lambda_{\mathrm{mono}}$ & $\lambda_{\mathrm{cont}}$ & Dev B4$\uparrow$ &
$\lambda_{\mathrm{mono}}$ & $\lambda_{\mathrm{cont}}$ & Dev B4$\uparrow$ \\
\midrule
0    & 0.2 & 26.10 & 0.1 & 0    & 27.10 \\
0.01 & 0.2 & 26.55 & 0.1 & 0.05 & 27.35 \\
0.05 & 0.2 & 27.10 & 0.1 & 0.1  & 27.22 \\
\textbf{0.1}  & \textbf{0.2} & \textbf{27.49} &
\textbf{0.1}  & \textbf{0.2} & \textbf{27.49} \\
0.2  & 0.2 & 27.25 & 0.1 & 0.4  & 26.64 \\
0.5  & 0.2 & 26.50 &     &      &       \\
1.0  & 0.2 & 25.95 &     &      &       \\
\bottomrule
\end{tabular}
\caption{Development-set sensitivity to the structural regularization weights. We report one-dimensional sweeps by varying one coefficient and fixing the other to its selected default. The best setting is $\lambda_{\mathrm{mono}}=0.1$ and $\lambda_{\mathrm{cont}}=0.2$ (bold).}
\label{tab:reg_sweep}
\end{table}


\begin{figure}[htbp]
    \centering
    \includegraphics[width=1\linewidth]{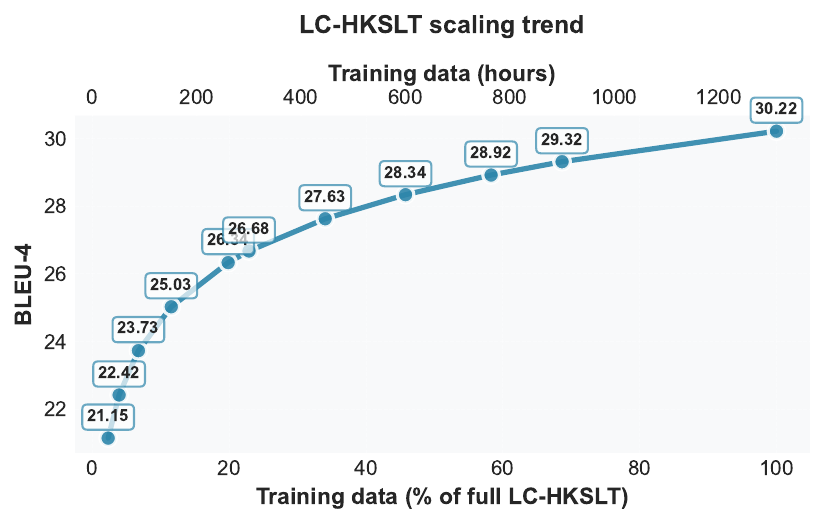}
    \caption{LC-HKSLT scaling trend.}
    \label{fig:lchk_scaling}
\end{figure}

\section{More Details on LC-HKSLT: Scaling, Quality, and Ethics}
\subsection{LC-HKSLT Dataset}
\label{app:lchkslt}
\begin{table*}[t]
\centering
\resizebox{1\linewidth}{!}{\begin{tabular}{l|c|c|c|c|c|c|c}
\toprule
Dataset / Work & Lang. & SLR Vocab. & SLT Vocab. & Duration (h) & Videos & Signers & Source \\
\midrule
GSL~\cite{adaloglou2021comprehensive}                 & GSL      & 310    & --      & 9.59 & 10K   & 7  & Lab \\
PHOENIX\textendash 14T \cite{camgoz2018neural}   & DGS      & 1{,}066& 2{,}887  & 10.53 & 8K    & 9  & TV  \\
CSL\textendash Daily \cite{zhou2021improving}   & CSL      & 2{,}000& 2{,}370 & 23.27 & 21K   & 10 & Lab \\
TVB\textendash HKSL\textendash News \cite{niu2024hongkongsignlanguage} & HKSL & 6{,}515 & 2{,}850 & 16.07& 7K & 2 & TV \\
\midrule
\textbf{LC\textendash HKSLT (Ours)} & HKSL     & --     & 125{,}833 & 1311& 432K & 14 & YouTube \\
\bottomrule
\end{tabular}}
\caption{Comparison of LC\textendash HKSLT with representative SLR/SLT datasets.}
\vspace{-1em}
\label{tab:hk_dataset_compare}
\end{table*}

\indent Our objective is to enable scalable, gloss-free sign language translation suitable for real-world deployment, where sign–text pairs are readily available but token-level annotations, such as glosses, are scarce. Existing Hong Kong Sign Language resources are typically limited in scale or rely heavily on gloss supervision, which is misaligned with learning directly from raw videos and sentence-level translations. To bridge this gap and support our Latent Chain-of-Thought framework for long-range, weakly supervised sign-to-text reasoning, we construct LC-HKSLT, a large-scale corpus curated from realistic broadcast-style scenarios.

\noindent\textbf{Collection pipeline.} We collect approximately 1,300 hours of Hong Kong Sign Language videos from YouTube, primarily drawn from Hong Kong Government and Legislative Council briefings in which an interpreter is continuously visible and the spoken content is publicly accessible. The videos are segmented into sentence-level clips, from which frames are extracted, and the corresponding speech tracks are transcribed using \texttt{openai/whisper-large-v3} to produce sentence-level targets. Notably, no gloss annotations or sign language recognition vocabularies are introduced. The resulting supervision consists solely of weak, sentence-level signals that are readily obtainable at scale in real-world settings, thereby aligning with the intended operating regime of our gloss-free formulation.

\noindent\textbf{Evaluation protocol.} Although LC-HKSLT is collected at web scale, this paper focuses on a carefully curated 30-hour subset, with zero sentence overlap across the training, validation, and test splits. This design choice is not intended to underrepresent the dataset, but rather to isolate the methodological contribution in a controlled, comparable setting. Specifically, (i) most existing Chinese SLT benchmarks operate in the 20 to 30 hour regime, enabling fair and direct comparison; (ii) the reduced scale allows training to remain practical on a single GPU; and (iii) we observe that, even without external pretraining, this data budget is sufficient to validate the core benefits of explicit latent reasoning. Table~\ref{tab:hk_dataset_compare} situates LC-HKSLT among representative SLR and SLT datasets. Beyond the subset used in this study, the complete collection comprises 432k clips from 14 signers and covers 125,833 Chinese words and phrases in the translation vocabulary. The full dataset will be publicly released to support future research on large-scale, gloss-free SLT. Beyond direct translation, large-scale sign-text corpora may also support retrieval-oriented multimodal settings that require composing linguistic intent with fine-grained visual evidence~\cite{HABIT}.

\subsection{Scaling Study on LC-HKSLT}
\label{app:lchk_scaling}
To address concerns about the gap between our web-scale collection and the controlled 30-hour evaluation protocol, we conduct an additional scaling study on LC-HKSLT by training the same model with increasing fractions of the training pool while keeping the dev/test split fixed.
As shown in Fig.~\ref{fig:lchk_scaling}, performance improves monotonically with more training data: BLEU-4 increases from 21.15 in the smallest-data setting to 30.22 when using the full LC-HKSLT collection (up to $\sim$1.3k hours), yielding a total gain of +9.07 BLEU-4.
Notably, improvements are most pronounced in the low-data regime and gradually taper off as the data scale grows, suggesting that our method is data-efficient while continuing to benefit from additional weakly supervised sign-text pairs.

\subsection{Data Quality, Licensing, and Privacy Considerations}
\label{app:lchk_ethics}
\paragraph{Data source and intended use.}
LC-HKSLT is collected from publicly accessible broadcast-style videos on YouTube, focusing on Hong Kong Government and Legislative Council (LegCo) briefings where a sign language interpreter is consistently visible and the spoken content is publicly available.
We segment videos into sentence-level clips and obtain sentence targets by transcribing the accompanying speech track using \texttt{openai/whisper-large-v3}. The dataset is intended solely for \textbf{non-commercial research and education} on sign language translation and related multimodal learning problems. More broadly, this line of work is also related to structured prediction over complex inputs, where multi-level semantic dependencies are important~\cite{zhang2025graphatc}.

\vspace{-0.5em}
\paragraph{Label noise and alignment limitations.}
Since sentence targets are derived from automatic speech recognition (ASR), LC-HKSLT may contain: (i) ASR transcription errors, (ii) sentence boundary/segmentation errors, and (iii) imperfect temporal alignment between the interpreter’s signing and the spoken content. Such issues are related to the broader challenge of learning under noisy supervision in text-centered settings~\cite{liu2023retrieval, liu2025queries,INTENT}.
We explicitly view LC-HKSLT as weak, sentence-level supervision at scale, which matches the real-world operating condition targeted by gloss-free SLT. We will release meta-information (e.g., clip timestamps and source video identifiers) to facilitate error analysis and downstream filtering by the community. Efficient hybrid nearest-neighbor search may further support scalable indexing and community-driven filtering over large clip collections~\cite{STABLE,ni2025unigeoseg}.

\vspace{-0.5em}
\paragraph{Privacy protection.}
Although the source videos are publicly available, we treat privacy as a first-class concern.
To mitigate re-identification risks in data release, we will:
(1) publish a privacy-preserving version of the clips where human faces are blurred (including the interpreter and any other visible individuals);
(2) avoid releasing personal identifiers beyond what is necessary for research (e.g., no names, no user handles), and provide only minimal source references (video ID and time range) for traceability;
(3) explicitly prohibit any attempt to use LC-HKSLT for biometric identification, surveillance, or re-identification.

\paragraph{Copyright and licensing considerations.}
LC-HKSLT is derived from third-party online videos. We do not claim ownership over the original content. Our release will follow a research-only license and will include clear attribution to the original sources, in line with broader concerns about unauthorized reuse of externally sourced data~\cite{liu2025stole}. If any source content owner or appearing individual requests removal, we will promptly honor such requests.

\paragraph{Take-down policy.}
We will maintain a take-down mechanism: upon receiving a request with the source video identifier and timestamp range, we will remove the corresponding clips/annotations from future releases and provide an updated index to the community.

\section{Additional Quantitative Results}
Fig.~\ref{fig:qualitative_vis} visualizes the frame-level thought-to-time alignment of our latent reasoning module on five CSL-Daily examples. For each video, we show: (i) the attention heatmap from $K{=}8$ latent thoughts ($C_1{\sim}C_8$) to input frames, together with the selectivity entropy $H$ and monotonic-violation rate \texttt{viol}; and (ii) the corresponding Chinese prediction with its English translation. The full model exhibits clear ordered bands (near-diagonal alignment), low \texttt{viol}, and relatively low $H$, indicating that different thoughts attend to progressively later evidence in a coherent manner. This observation is broadly consistent with the importance of spatially aligned and semantically discriminative representations in complex visual recognition settings~\cite{yao2025s,zhang2025high}. In contrast, removing $\mathcal{L}_{mono}$ disrupts temporal ordering (higher \texttt{viol}) and often leads to semantic reversals or role/polarity mistakes (e.g., predicting “can tell the difference” instead of “cannot”). Removing $\mathcal{L}_{cont}$ fragments attention across disjoint regions, correlating with disfluent or incomplete outputs. Replacing structured routing with soft routing produces diffuse, high-entropy allocations and more generic translations that miss key details. Overall, these cases qualitatively support that our chain-structured, regularized evidence allocation yields more faithful and grounded translations, especially for longer sentences requiring multi-step semantic composition. These visual patterns correspond closely to linguistic coherence in the generated translations: without the regularizers, models tend to swap event order or omit key arguments, confirming that chain-structured reasoning improves both alignment and semantic faithfulness~\cite{jiang2026foeforesterrorsmakes}.

\end{document}